\documentclass[a4paper,10pt]{article}
\usepackage[utf8]{inputenc}
\usepackage[noadjust]{cite}
\usepackage{amssymb,amsmath}
\usepackage[margin=0.75in]{geometry}
\usepackage{xcolor}

\usepackage{amsmath,amssymb,amsthm}

\newtheorem{cor}{Corollary}
\theoremstyle{definition}

\title{Class-Conditional Compression and Disentanglement: Bridging the Gap between Neural Networks and Naive Bayes Classifiers}
\author{Rana Ali Amjad and Bernhard C. Geiger\footnote{The authors contributed equally to this document.}}
\date{\today}

\newcommand{\ent}[1]{H(#1)}

\newcommand{\mutinf}[1]{I(#1)}
\newcommand{\kl}[2]{D\left(#1\|#2\right)}

\newcommand{\expop}[2]{E_{#1}\left[#2\right]}

\newcommand{\TC}[1]{TC(#1)}

\newcommand{\pmf}[1]{p_{#1}}
\newcommand{\smap}[1]{q_{#1}}
\newcommand{\surrdist}[1]{r_{#1}}

\newcommand{\dom}[1]{\mathcal{#1}}

\newcommand{\costib}{\dom{L}_\mathrm{IB}}
\newcommand{\costcib}{\dom{L}_\mathrm{CIB}}

\newcommand{\comm}[1]{}
\begin{document}

\maketitle

\begin{abstract}
 In this draft, which reports on work in progress, we 1) adapt the information bottleneck functional by replacing the compression term by class-conditional compression, 2) relax this functional using a variational bound related to class-conditional disentanglement, 3) consider this functional as a training objective for stochastic neural networks, and 4) show that the latent representations are learned such that they can be used in a naive Bayes classifier. We continue by suggesting a series of experiments along the lines of Nonlinear Information Bottleneck~\cite{Kolchinsky_NLIB}, Deep Variational Information Bottleneck~\cite{Alemi_DVIB}, and Information Dropout~\cite{Achille_InfoDropout}. We furthermore suggest a neural network where the decoder architecture is a parameterized naive Bayes decoder.
\end{abstract}

\section{Notation}\label{sec:notation}
We consider a classification task with a feature random variable (RV) $X$ on $\mathbb{R}^m$ and a class RV $Y$ on the finite set $\dom{Y}$ of classes. If a dataset $\dom{D}$ is available, then this dataset consists of $N$ realizations $(x_i,y_i)$ of the joint distribution $\pmf{XY}$, i.e., $\dom{D}=\{(x_i,y_i)\}_{i=1,\dots,N}$.

We further consider stochastic feed-forward neural networks (NNs). We assume that the input of the NN is the RV $X$, the output of the network is the RV $\hat{Y}$, and every hidden layer defines an internal representation. In this work we are interested in a particular representation at a dedicated bottleneck layer, which we will denote by the RV $T$. The NN is parameterized by a set $\Theta$ of parameters which define the stochastic map $\smap{T|X}$ from the input to the representation and the stochastic map $\smap{\hat{Y}|T}$ from the representation to the network output. We call $\smap{T|X}$ and $\smap{\hat{Y}|T}$ the encoder and decoder, respectively.

With this notation established, we denote distributions that are induced by the encoder/decoder (i.e., that depend on the parameters $\Theta$) with $q$. For example, we have 
\begin{equation}
\smap{Y|T}(y|t):=\frac{\expop{X\sim\pmf{X|Y}(\cdot|y)}{\smap{T|X}(t|X)}}{\expop{X\sim\pmf{X}}{\smap{T|X}(t|X)}}
\end{equation}
for the distribution of the class label conditioned on the latent representation and
\begin{equation}
 \smap{T|Y}(t|y):=\expop{X\sim\pmf{X|Y}(X|y)}{\smap{T|X}(t|X)}
\end{equation}
for the distribution of the latent representation conditioned on the class label. Surrogate distributions are denoted with $r$.

\section{Adapting the Information Bottleneck Loss for Optimal Representations}\label{sec:IB}
Our aim is to extract a representation $T$ of the feature $X$ such that the representation allows an accurate classification, but that at the same time is maximally compressed. In other words, we are looking for a stochastic map of $X$ such that the output $T$ of this map contains all -- but not more -- information about the class $Y$ that is contained in $X$. This aim is often formalized in terms of the information bottleneck (IB) functional; in the notation of~\cite[eq.~(2)]{Achille_InfoDropout}, we aim to find a minimizer $\pmf{T|X}$ of
\begin{equation}\label{eq:IB}
 \costib := \ent{Y|T} + \beta\mutinf{X;T}
\end{equation}
where $\beta\in[0,1]$ trades between the aims of preserving information about $Y$ (first term) and compressing the representation $T$ (second term). These two goals are conflicting, because compressing the representation potentially also leads to a loss of information relevant for classification. In the extreme case where $\mutinf{X;T}=0$ we trivially have $\ent{Y|T}=\ent{Y}$.

We now show that a different but equivalent formulation of the IB functional leads to two terms which not in direct conflict anymore. Specifically, we replace the compression term by a class-conditional compression term: Our aim is not to compress the latent representation $T$, but to remove every bit of information from this latent representation that is not necessary for classification. This latter quantity is captured in the conditional mutual information $\mutinf{X;T|Y}$.

Indeed, since $Y-X-T$ is a Markov tuple, we have that $\mutinf{X;T}=\mutinf{X,Y;T}$. Furthermore, by the chain rule of mutual information, we have
\begin{equation}\label{eq:chainrule}
 \mutinf{X;T}=\mutinf{X;T|Y}+\mutinf{Y;T} = \mutinf{X;T|Y}+\ent{Y}-\ent{Y|T}.
\end{equation}
Inserting~\eqref{eq:chainrule} into~\eqref{eq:IB} yields
\begin{align}
 \costib &= \ent{Y|T} + \beta\mutinf{X;T|Y}+\beta\ent{Y}-\beta\ent{Y|T}\notag\\
 &= (1-\beta)\ent{Y|T} + \beta\mutinf{X;T|Y}+\beta\ent{Y}
\end{align}
Since $\ent{Y}$ is independent of the map $\pmf{T|X}$, $\pmf{T|X}$ minimizes $\costib$ for $\beta\in[0,1]$ if and only if it minimizes
\begin{equation}\label{eq:newIB}
 \costcib := \ent{Y|T} + \beta'\mutinf{X;T|Y}
\end{equation}
for $\beta'=\beta/(1-\beta)$. Minimizing the second term -- which we call \emph{class-conditional compression} in the remainder of this work -- is not in direct conflict with minimizing the first anymore, as $\mutinf{X;T|Y}=0$ and $\ent{Y|T}=0$ are jointly possible.\footnote{Going one step further, noticing that $\ent{Y|X}$ does not depend on $\pmf{T|X}$, and that $\mutinf{X;Y|T}=\ent{Y|T}-\ent{Y|X}$, one can show that the optimization problem is equivalent to finding a minimizer of
\begin{equation}
 \mutinf{X;Y|T} + \beta'\mutinf{X;T|Y}
\end{equation}
for some $\beta\ge0$. The first term is a measure of sufficiency of the representation $T$~\cite[Sec.~4]{Achille_InfoDropout}, while the second term quantifies whether the representation is minimal in the sense of removing \emph{irrelevant} information.}

Taking a closer look at the fact that $\beta'=\beta/(1-\beta)$ illustrates that for $\beta\to 1$ we have $\beta'\to\infty$, i.e., the optimization problem focuses only on (class-conditional) compression. This has been observed both analytically (e.g.,~\cite[p.~2]{Kolchinsky_NLIB}) and in experiments (e.g.,~\cite[Figs.~4~\&~5]{Achille_InfoDropout} and~\cite[Fig.~1]{Alemi_DVIB}).

\section{A Variational Bound on Class-Conditional Compression and Its Consequences}
While we have shown that the functional~\eqref{eq:IB}, and thus also~\eqref{eq:newIB} becomes infinite for deterministic NNs with a continuously distributed input~\cite[Th.~1]{Amjad_LearningRepresentations}, for stochastic NNs it was argued that these functionals are complicated to estimate~\cite{Kolchinsky_NLIB,Alemi_DVIB}. As a remedy, both terms of $\costib$ can be replaced by variational bounds. We aim to do the same here for $\costcib$.

We start with $\ent{Y|T}$:
\begin{align}
 \ent{Y|T} &= \expop{X,Y\sim\pmf{XY}}{\expop{T\sim\smap{T|X}}{-\log\smap{Y|T}(Y|T)}}\\
 &=\expop{X,Y\sim\pmf{XY}}{\expop{T\sim\smap{T|X}}{-\log\smap{\hat{Y}|T}(Y|T)}} - \expop{X,T\sim\pmf{X}\smap{T|X}}{\kl{\smap{Y|T}(\cdot|T)}{\smap{\hat{Y}|T}(\cdot|T)}}\\
 &\le \expop{X,Y\sim\pmf{XY}}{\expop{T\sim\smap{T|X}}{-\log\smap{\hat{Y}|T}(Y|T)}}
\end{align}
where the inequality follows from the non-negativitiy of KL divergence and leads to the popular cross-entropy cost function. For the second term $\mutinf{X;T|Y}$, note that by the non-negativity of KL divergence we have
\begin{subequations}\label{eq:classcondderiv}
 \begin{align}
\mutinf{X;T|Y} &= \expop{X,Y\sim\pmf{XY}}{\expop{T\sim\smap{T|X}}{\log\frac{\smap{T|X}(T|X)}{\smap{T|Y}(T|Y)}}}\\
&= \expop{X,Y\sim\pmf{XY}}{\expop{T\sim\smap{T|X}}{\log\frac{\smap{T|X}(T|X)}{\surrdist{T|Y}(T|Y)}}} - \expop{Y\sim\pmf{Y}}{\kl{\smap{T|Y}(\cdot|Y)}{\surrdist{T|Y}(\cdot|Y)}}\\
&\le \expop{X,Y\sim\pmf{XY}}{\expop{T\sim\smap{T|X}}{\log\frac{\smap{T|X}(T|X)}{\surrdist{T|Y}(T|Y)}}} \label{eq:variationalBound}\\
&=\expop{X,Y\sim\pmf{XY}}{ \kl{\smap{T|X}(\cdot|X)}{\surrdist{T|Y}(\cdot|Y)}}
\end{align}
\end{subequations}
for any surrogate distribution $\surrdist{T|Y}$. Combining both terms and evaluating the outer expectation by averaging over a dataset $\dom{D}$, we obtain the following cost function for NN training:
\begin{equation}\label{eq:newCost}
 \costcib^*(\dom{D}) := \frac{1}{N}\sum_{i=1}^N \expop{T\sim\smap{T|X}(\cdot|x_i)}{-\log\smap{\hat{Y}|T}(y_i|T)} + \beta' \kl{\smap{T|X}(\cdot|x_i)}{\surrdist{T|Y}(\cdot|y_i)}
\end{equation}
For a fixed $\surrdist{T|Y}$, this cost function is minimized over $\smap{\hat{Y}|T}$ and $\smap{T|X}$, or equivalently, over the parameters of the NN. More generally, if $\surrdist{T|Y}$ can be selected from a family of distributions, then $\costcib^*(\dom{D})$ is minimized over the parameters of the NN and over all $\surrdist{T|Y}$ within this family.

Since~\eqref{eq:variationalBound} holds for every surrogate distribution, it also holds for a product distribution over the components of $T$, i.e., for $\surrdist{T|Y}=\surrdist{T}=\prod\surrdist{T_j}$, where $T_j$ is the $j$-th neuron in the bottleneck layer. This choice yields the variational bounds in~\cite{Alemi_DVIB,Achille_InfoDropout}. In contrast, we make the assumption that the distribution of the representation $T$ factorizes when conditioning on the class variable $Y$. In other words, we set 
\begin{equation}\label{eq:NB}
\surrdist{T|Y}=\prod\surrdist{T_j|Y}. 
\end{equation}

In a generative auto-encoding setup in which no class labels are present (or even meaningful), the setting 
$\surrdist{T|Y}=\prod\surrdist{T_j}$ makes sense: Generating a sample of $X$ amounts to sampling from $\surrdist{T}$, which is particularly simple if the components of $T$ are independent.\footnote{The authors of~\cite{Achille_InfoDropout} build a connection between information dropout and variational auto-encoders (VAE)~\cite{Kingma_VAE}. Specifically, they argue that the variational bound on $\costib$ corresponding to $\costcib^*$ is equivalent to the cost function of the VAE when $\beta=1$. We wish to note here that the IB functional $\costib$ itself is not meaningful in an auto-encoding setup, i.e., for $Y\equiv X$: In this case, we have $\costib=\ent{X}$ for $\beta=1$, i.e., the cost is independent of the encoder $\smap{T|X}$ and the decoder $\smap{X|T}$. For $\beta<1$, the IB functional aims at minimizing $\ent{X|T}$, which is trivially fulfilled by an encoder that makes $T$ independent of $X$. Auto-encoding as a trade-off between compression and reconstruction fidelity is only obtained after bounding $\ent{X|T}$ with the cross-entropy induced by the decoder distribution.} As soon as class labels are available, we argue that~\eqref{eq:NB} is preferable over the unconditional setting $\surrdist{T|Y}=\prod\surrdist{T_j}$. This is obvious for the classification task; e.g., it is easier to build a classifier operating on a Gaussian mixture model than on a Gaussian RV, cf. Section~\ref{sec:NB}. 

However, even for a generative auto-encoding setup,~\eqref{eq:NB} makes sense if class labels are available. In this case, the aim of the decoder is to reconstruct the input $X$ from the latent representation $T$, i.e., the decoder has the structure $\smap{\hat{X}|T}$. Generating an example of a given class $y$ amounts to sampling from $\surrdist{T|Y}(\cdot|y)$, i.e., the distribution over which one samples depends on the class of which one wants to generate an example. (And sampling from this distribution is particularly simple if is a product distribution.) This conditional variational auto-encoding (CVAE) was discussed in~\cite{Sohn_CVAE} for the case where both encoder and decoder may depend on the class variable, i.e., for $\smap{T|X,Y}$ and $\smap{\hat{X}|T,Y}$. Removing this dependences on the class variable, their cost function~\cite[eq.~(4)]{Sohn_CVAE} is equivalent to our~\eqref{eq:newCost} for $\beta'=1$ and for $\smap{\hat{Y}|T}$ exchanged with $\smap{\hat{X}|T}$.

\subsection{First Consequence: Naive Bayes Structure}\label{sec:NB}
An immediate consequence of~\eqref{eq:classcondderiv} is that minimizing~\eqref{eq:newCost} for~\eqref{eq:NB} simultaneously encourages an encoder $\smap{T|X}$ that leads to class-conditional compression and a naive Bayes structure that can be exploited by the decoder. This follows because
\begin{equation}\label{eq:decomposition}
 \expop{X,Y\sim\pmf{XY}}{ \kl{\smap{T|X}(\cdot|X)}{\prod\surrdist{T_j|Y}(\cdot|Y)}} = \mutinf{X;T|Y}+\expop{Y\sim\pmf{Y}}{\kl{\smap{T|Y}(\cdot|Y)}{\prod\surrdist{T_j|Y}(\cdot|Y)}}.
\end{equation}
Specifically, suppose that the second term~\eqref{eq:decomposition} vanishes. Then, $\smap{T|Y}=\prod\surrdist{T_j|Y}$ almost surely, and the optimal decoder $\smap{\hat{Y}|T}$ is a naive Bayes classifier.

From this perspective, the following approach seems to make sense: One fixes a family of distributions from which $\surrdist{T|Y}=\prod\surrdist{T_j|Y}$ is taken; e.g., $\surrdist{T|Y}$ could be a multivariate Gaussian distribution with mean vector and diagonal covariance matrix that depend on the class label. For this parameterized family of distributions, one fixes the decoder $\smap{\hat{Y}|T}$ to be the corresponding naive Bayes classifier. Then, by~\eqref{eq:decomposition}, minimizing~\eqref{eq:newCost} over the encoder $\smap{T|X}$ and the parameters of $\surrdist{T|Y}$ leads to an encoder network such that 1) the latent representations are such that the support a naive Bayes classifier, 2) the naive Bayes classifier has good performance on the latent representations, and 3) the latent representations are class-conditionally compressed.

\subsection{Second Consequence: Class-Conditional Disentanglement}\label{sec:CCDisent}
In the more general case in which $\costib(\dom{D})$ is minimized over the parameters of the NN and over all $\surrdist{T|Y}=\surrdist{T}$ within a given family, it was shown that~\cite[Proposition~1]{Achille_InfoDropout}
\begin{subequations}\label{eq:dropoutequivalence}
\begin{equation}
 \min_{\smap{T|X},\smap{\hat{Y}|T},\{\surrdist{T_j}\}} \frac{1}{N}\sum_{i=1}^N \expop{T\sim\smap{T|X}(\cdot|x_i)}{-\log\smap{\hat{Y}|T}(y_i|T)} + \beta \kl{\smap{T|X}(\cdot|x_i)}{\prod\surrdist{T_j}(\cdot)}
\end{equation}
is equivalent to
\begin{equation}
 \min_{\smap{T|X},\smap{\hat{Y}|T}} \frac{1}{N}\sum_{i=1}^N \expop{T\sim\smap{T|X}(\cdot|x_i)}{-\log\smap{\hat{Y}|T}(y_i|T)} + \beta \kl{\smap{T|X}(\cdot|x_i)}{\smap{T}(\cdot)} + \beta \TC{T}
\end{equation}
\end{subequations}
where $\TC{T}=\kl{\smap{T}}{\prod\smap{T_j}}$ is the total correlation and where $\smap{T}(t)=\frac{1}{N}\sum_{i=1}^N \smap{T|X}(t|x_i)$. In other words, minimizing $\costcib^*(\dom{D})$ for the setting $\surrdist{T|Y}=\prod\surrdist{T_j}$ encourages disentangled representations.

If instead of $\surrdist{T|Y}=\prod\surrdist{T_j}$ we set $\surrdist{T|Y}=\prod\surrdist{T_j|Y}$, then one can show that \emph{conditionally disentangled} representations are encouraged. In other words, the extracted features are not required to be independent, but to be conditionally independent given the class variable. We believe that this conditional disentanglement is theoretically preferable over disentanglement, if some kind of disentanglement is preferable at all.

\begin{cor}[Corollary to {\cite[Proposition~1]{Achille_InfoDropout}}]\label{cor:TC}
The minimization problem
 \begin{subequations}
\begin{equation}
 \min_{\smap{T|X},\smap{\hat{Y}|T},\{\surrdist{T_j|Y}\}} \frac{1}{N}\sum_{i=1}^N \expop{T\sim\smap{T|X}(\cdot|x_i)}{-\log\smap{\hat{Y}|T}(y_i|T)} + \beta \kl{\smap{T|X}(\cdot|x_i)}{\prod\surrdist{T_j|Y}(\cdot|y_i)}
\end{equation}
is equivalent to
\begin{equation}\label{eq:cor:second}
 \min_{\smap{T|X},\smap{\hat{Y}|T}} \frac{1}{N}\sum_{i=1}^N \expop{T\sim\smap{T|X}(\cdot|x_i)}{-\log\smap{\hat{Y}|T}(y_i|T)} + \beta \kl{\smap{T|X}(\cdot|x_i)}{\smap{T|Y}(\cdot|y_i)} + \beta \TC{T|y_i}
\end{equation}
\end{subequations}
where $\TC{T|y_i}=\kl{\smap{T|Y}(\cdot|y_i)}{\prod\smap{T_j|Y}(\cdot|y_i)}$ and 
$\smap{T|Y}(t|y)=\frac{1}{|\{i{:}\ y_i=y\}|}\sum_{i{:}\ y_i=y}\smap{T|X}(t|x_i)$.
\end{cor}

Before providing the proof, two aspects are worth mentioning. First, the equivalence of the two optimization problems in the corollary is only valid if the optimization over the marginal distributions $\{\surrdist{T_j|X}\}$ is unconstrained. If instead, for example, the distributions $\{\surrdist{T_j|Y}\}$ have to be chosen from a specific family (e.g., Gaussian), then this equivalence need not hold in general. We believe that such a constrained optimization is of greater practical relevance than the unconstrained one, which in some sense limits the practical applicability of this result. The second aspect is that, if instead of a dataset $\dom{D}$ the distribution $\pmf{X,Y}$ is used to compute expectations, the second and third terms in~\eqref{eq:cor:second} evaluate to $\mutinf{X;T|Y}$ and $\TC{T|Y}:=\sum_{j} \ent{T_j|Y}-\ent{T|Y}$. Thus, and connecting to~\eqref{eq:decomposition}, it can be seen that the variational bound on $\mutinf{X;T|Y}$ is equivalent to adding a regularization term that encourages disentanglement (cf. the discussion after~\cite[Proposition~1]{Achille_InfoDropout}).

\begin{proof}
  The first term does not depend on $\surrdist{T|Y}$, so it suffices to show that
  \begin{equation}
   \min_{\{\surrdist{T_j|Y}\}} \frac{1}{N}\sum_{i=1}^N\kl{\smap{T|X}(\cdot|x_i)}{\prod\surrdist{T_j|Y}(\cdot|y_i)} = \frac{1}{N}\sum_{i=1}^N \kl{\smap{T|X}(\cdot|x_i)}{\smap{T|Y}(\cdot|y_i)} + \TC{T|y_i}
  \end{equation}
  for every $\smap{T|X},\smap{\hat{Y}|T}$. Indeed, by the product rule of the logarithm one can show that
  \begin{align}
   &\frac{1}{N}\sum_{i=1}^N\kl{\smap{T|X}(\cdot|x_i)}{\prod\surrdist{T_j|Y}(\cdot|y_i)}\notag\\
   &=\frac{1}{N}\sum_{i=1}^N \expop{T\sim\smap{T|X}(\cdot|x_i)}{\log\frac{\smap{T|X}(T|x_i)}{\smap{T|Y}(T|y_i)}} + \expop{T\sim\smap{T|X}(\cdot|x_i)}{\log\frac{\smap{T|Y}(T|y_i)}{\prod\surrdist{T_j|Y}(T|y_i)}}\\
   &= \frac{1}{N}\sum_{i=1}^N \kl{\smap{T|X}(\cdot|x_i)}{\smap{T|Y}(\cdot|y_i)} +  \expop{T\sim\smap{T|X}(\cdot|x_i)}{\log\frac{\smap{T|Y}(T|y_i)}{\prod\surrdist{T_j|Y}(T|y_i)}}
  \end{align}
  It remains to show that minimizing the second part of this sum over all $\{\surrdist{T_j|Y}\}$ yields $\frac{1}{N}\sum_{i=1}^N\TC{T|y_i}$. To this end, we split the sum over all samples over two sums, one of which runs over the possible values $y$ of the class variable, and one that runs over all samples $(x_i,y_i)$ for which $y_i=y$. With this, and the law of total expectation, we get
  \begin{align}
   \frac{1}{N}\sum_{i=1}^N \expop{T\sim\smap{T|X}(\cdot|x_i)}{\log\frac{\smap{T|Y}(T|y_i)}{\prod\surrdist{T_j|Y}(T|y_i)}}
   &=\frac{1}{N}\sum_{y\in\dom{Y}}\sum_{i{:}\ y_i=y}\expop{T\sim\smap{T|X}(\cdot|x_i)}{\log\frac{\smap{T|Y}(T|y)}{\prod\surrdist{T_j|Y}(T|y)}}\\
   &=\frac{1}{N}\sum_{y\in\dom{Y}} |\{i{:}\ y_i=y\}| \expop{T\sim\smap{T|Y}(\cdot|y)}{\log\frac{\smap{T|Y}(T|y)}{\prod\surrdist{T_j|Y}(T|y)}}\\
   &=\frac{1}{N}\sum_{y\in\dom{Y}} |\{i{:}\ y_i=y\}| \kl{\smap{T|Y}(\cdot|y)}{\prod\surrdist{T_j|Y}(\cdot|y)}\label{eq:proof:tomin}
  \end{align}
  We now minimize the right-hand side of~\eqref{eq:proof:tomin} over all $\{\surrdist{T_j|Y}\}$. To this end, for every $y$, we expand the KL divergence via the chain rule~\cite[Th.~2.5.3]{Cover_Information} to get
  \begin{align}
   &\min_{\{\surrdist{T_j|Y}\}}\frac{1}{N}\sum_{y\in\dom{Y}} |\{i{:}\ y_i=y\}| \kl{\smap{T|Y}(\cdot|y)}{\prod\surrdist{T_j|Y}(\cdot|y)} \notag\\
   &= \frac{1}{N}\sum_{y\in\dom{Y}} |\{i{:}\ y_i=y\}| \sum_j \min_{\surrdist{T_j|Y}}\expop{T_1^{j-1}\sim\smap{T_1^{j-1}|Y}(\cdot|y)}{ \kl{\smap{T_j|Y,T_1^{j-1}}(\cdot|y,T_1^{j-1})}{\surrdist{T_j|Y}(\cdot|y)}}\\
   &\stackrel{(a)}{=} \frac{1}{N}\sum_{y\in\dom{Y}} |\{i{:}\ y_i=y\}| \sum_j \expop{T_1^{j-1}\sim\smap{T_1^{j-1}|Y}(\cdot|y)}{ \kl{\smap{T_j|Y,T_1^{j-1}}(\cdot|y,T_1^{j-1})}{\smap{T_j|Y}(\cdot|y)}}\\
   &=\frac{1}{N}\sum_{y\in\dom{Y}} |\{i{:}\ y_i=y\}| \kl{\smap{T|Y}(\cdot|y)}{\prod\smap{T_j|Y}(\cdot|y)} = \frac{1}{N}\sum_{i=1}^N\TC{T|y_i}
  \end{align}
  where $(a)$ follows from~\cite[Lemma~13.8.1]{Cover_Information}. This completes the proof.
\end{proof}

\section{Planned Experiments}
To investigate whether the presented framework based on class-conditional compression is useful, we plan to perform a set of experiments. Whether these experiments are feasible in principle is, at present, unclear.

\subsection{Nonlinear Information Bottleneck}
In~\cite{Kolchinsky_NLIB}, the authors use a stochastic encoder $\smap{T|X}$ which learns the mean vector of a multivariate Gaussian with identity covariance matrixm, i.e., $\smap{T|X}(\cdot|x)\sim\dom{N}(f_\theta(x),\sigma^2\mathbf{I})$. Therefore, the authors assume that the latent representation $T$ is a Gaussian mixture, with each point in the dataset being an individual component. Based on this assumption, they propose bounding the compression term via~\cite[eq.~(10)]{Kolchinsky_NLIB}
\begin{equation}\label{eq:NIB}
 \mutinf{X;T} \le -\frac{1}{N}\sum_{i=1}^N \log \sum_{j=1}^N \exp{\left(-\frac{1}{2}\frac{\|f_\theta(x_i)-f_\theta(x_j)\|}{\eta^2(\theta)+\sigma^2}\right)}-m\log\frac{\sigma^2}{\eta^2(\theta)+\sigma^2}
\end{equation}
where $\eta(\theta)$ is a noise parameter that is learned.

Moving from compression to class-conditional compression is achieved by replacing $\mutinf{X;T}$ by $\mutinf{X;T|Y}$. We believe that this should also be possible in the framework of nonlinear information bottleneck by computing~\eqref{eq:NIB} separately for each class. In other words, we bound
\begin{equation}
 \mutinf{X;T|Y=y} \le -\frac{1}{N}\sum_{i{:}\ y_i=y} \log \sum_{i{:}\ y_i=y} \exp{\left(-\frac{1}{2}\frac{\|f_\theta(x_i)-f_\theta(x_j)\|}{\eta^2(\theta)+\sigma^2}\right)}-m\log\frac{\sigma^2}{\eta^2(\theta)+\sigma^2} =: \hat{I}(X;T|Y=y)
\end{equation}
and obtain
\begin{equation}
 \mutinf{X;T|Y} \le \sum_{y\in\dom{Y}} |\{i{:}\ y_i=y\}| \hat{I}(X;T|Y=y).
\end{equation}

\subsection{Naive Bayes Decoder}
This experiment is based on Section~\ref{sec:NB}. Specifically, we plan to choose $\surrdist{T|Y}$ from the family of Gaussian distributions with a mean vector $\mu_y$ that depends on the class $y$ and an identity matrix (possibly scaled with a constant $\sigma_y$ that depends on the class $y$) as covariance matrix. This leads to the goal of obtaining a latent representation $T$ that is well-approximated by a Gaussian mixture model, where each mixture component is spherical.

Rather than training the decoder part $\smap{\hat{Y}|T}$ of the network, we replace this part by a naive Bayes classifier fitted to the parameters $\{\mu_y,\sigma_y\}$ of $\surrdist{T|Y}$. Our aim is then to train the encoder part of the network such that the naive Bayes decoder can be fully exploited, i.e., we learn the parameters of the encoder and the parameters $\{\mu_y,\sigma_y\}$ of $\surrdist{T|Y}$ such that cost $\costcib^*(\dom{D})$ is minimized.

\subsection{Deep Variational Information Bottleneck}
The authors of~\cite{Alemi_DVIB} suggest minimizing $\costcib^*$ for a spherical Gaussian $\surrdist{T}$, i.e., they assume that $\surrdist{T}\sim\dom{N}(0,\mathbf{I})$. Replacing this target distribution by a conditionally independent distribution of the latent dimensions given the class, i.e., by $\surrdist{T|Y}(\cdot|y)\sim\dom{N}(\mu_y,\mathbf{I})$ is simple. Unclear is, how the mean vectors $\{\mu_y\}$ shall be chosen or -- which is preferable in the light of Corollary~\ref{cor:TC} -- if these mean vectors can be learned from data jointly (or alternatingly) with the remaining network parameters.

\subsection{Conditional Information Dropout}
In~\cite{Achille_InfoDropout}, the authors made the connection between a well-chosen variational bound and disentanglement~\cite[Proposition~1]{Achille_InfoDropout}. They further proposed an encoder $\smap{T|X}$ that is implemented by a NN where each neuron output is affected by multiplicative data-dependent noise (which is chosen to follow a log-normal distribution with data-dependent variance for the sake of analytical simplicity). The authors furthermore proposed that $\surrdist{T}=\prod\surrdist{T_i}$, where $\surrdist{T_i}$ is log-uniform with a point mass at zero or log-normal for ReLU or softplus activation functions, respectively (cf.~\cite[Propositions~2~and~3]{Achille_InfoDropout}).

In the setting proposed in this draft, one would have to replace $\prod\surrdist{T_i}$ by $\prod\surrdist{T_i|Y}$. In case of a softplus activation, this would mean that $\surrdist{T_i|Y}$ is a log-normal distribution the mean of which depends on the class (and potentially on the latent dimension $i$). In case of a ReLU activation, this would require that the point mass at zero depends on the class (and potentially on the latent dimension $i$). We are again faced with the issued mentioned in the previous subsection, i.e., whether these parameters can be trained from data or if (and how) they can be selected a priori.

\bibliography{../references.bib}
\bibliographystyle{apalike}

\end{document}